\documentclass[letterpaper, 10 pt, conference]{ieeeconf}  % Comment this line out if you need a4paper

\IEEEoverridecommandlockouts                              % This command is only needed if 
\usepackage{graphics} % for pdf, bitmapped graphics files
\usepackage{epsfig} % for postscript graphics files
\usepackage{mathptmx} % assumes new font selection scheme installed
\usepackage{times} % assumes new font selection scheme installed
\usepackage{amsmath} % assumes amsmath package installed
\usepackage{amssymb}  % assumes amsmath package installed

\usepackage{booktabs}       % professional-quality tables
\usepackage{array}

\usepackage{amsfonts}       % blackboard math symbols
\usepackage{xcolor}         % color
\title{\LARGE \bf
Mem2Ego: Empowering Vision-Language Models with Global-to-Ego Memory for Long-Horizon Embodied Navigation
}

\author{Lingfeng Zhang$^{1}$, Yuecheng Liu$^{1}$, Zhanguang Zhang$^{1}$, Matin Aghaei$^{1}$, Yixin Xiao$^{1}$, Yaochen Hu$^{1}$, \\
Mohammad Ali Alomrani$^{1}$, David Gamaliel Arcos Bravo$^{1}$, Hongjian Gu$^{1}$, Zhiyuan Li$^{1}$, Yangzheng Wu$^{1}$, \\
Zhanpeng Zhang$^{1}$, Raika Karimi$^{1}$, Atia Hamidizadeh$^{1}$, Guowei Huang$^{1}$, Haoping Xu$^{1}$, Tongtong Cao$^{1}$, \\
Weichao Qiu$^{1}$, Xingyue Quan$^{1}$, Jianye Hao$^{1}$, Yuzheng Zhuang$^{1}$, Yingxue Zhang$^{1}$
\thanks{$^{1}$Huawei Noah's Ark Lab {\tt\small lingfeng.zhang1@huawei.com, zhanguang.zhang@huawei.com, yingxue.zhang@huawei.com}}%
}

\begin{document}

\maketitle
\thispagestyle{empty}
\pagestyle{empty}

%%%%%%%%%%%%%%%%%%%%%%%%%%%%%%%%%%%%%%%%%%%%%%%%%%%%%%%%%%%%%%%%%%%%%%%%%%%%%%%%

\begin{abstract}
Recent advancements in Large Language Models (LLMs) and Vision-Language Models (VLMs) have made them powerful tools in embodied navigation, enabling agents to leverage commonsense and spatial reasoning for efficient exploration in unfamiliar environments. Existing LLM-based approaches convert global memory, such as semantic or topological maps, into language descriptions to guide navigation. While this improves efficiency and reduces redundant exploration, the loss of geometric information in language-based representations hinders spatial reasoning, especially in intricate environments. To address this, VLM-based approaches directly process ego-centric visual inputs to select optimal directions for exploration. However, relying solely on a first-person perspective makes navigation a partially observed decision-making problem, leading to suboptimal decisions in complex environments.
In this paper, we present a novel vision-language model (VLM)-based navigation framework that addresses these challenges by adaptively retrieving task-relevant cues from a global memory module and integrating them with the agent’s egocentric observations. By dynamically aligning global contextual information with local perception, our approach enhances spatial reasoning and decision-making in long-horizon tasks. 
% Experimental results demonstrate that the proposed method surpasses previous state-of-the-art approaches in object navigation tasks, providing a more effective and scalable solution for embodied navigation.
The proposed method surpasses previous state-of-the-art approaches by a significant margin on both the HSSD and HM3D benchmarks and demonstrates strong performance on a real robot.
\end{abstract}

\section{Introduction}
Embodied navigation is a crucial component of embodied artificial intelligence, with widespread applications in diverse scenarios such as  domestic environments, office settings, logistics delivery, and factory inspections \cite{savva2019habitat, duan2022survey, zhang2022survey}. Its significance stems from its ability to enable agents to autonomously navigate and perform tasks within physical environments \cite{liu2024aligning, moller2021survey}.

Embodied navigation poses two key challenges. First, unlike autonomous driving, which typically occurs in structured outdoor environments, embodied navigation requires operating in diverse indoor and industrial settings such as factories, shopping malls, and offices. These spaces feature intricate layouts and obstacles, demanding advanced perception and planning \cite{thorisson2012cognitive, salvini2022safety, duan2022survey, mavrogiannis2022social}. Second, it necessitates a high degree of autonomy, as agents must adapt to unfamiliar environments without relying on predefined maps. They must interpret human instructions and dynamically interact with their surroundings to navigate effectively. This work focuses on Object Goal Navigation (ObjectNav), a task in which agents must locate specified objects within complex spaces \cite{batra2020objectnav, majumdar2022zson}.

In recent years, the rapid advancement of large language models (LLMs) has opened new possibilities for embodied navigation~\cite{zeng2023large, wu2024embodied}. These models enable robots to leverage commonsense reasoning, improving their understanding of natural language commands and enhancing the integration of perceptual data. This allows for navigation decisions that better align with human intentions \cite{wu2024voronav, li2024embodied}. Furthermore, recent ObjectNav research underscores the importance of historical information in improving environmental understanding, decision-making, and grounding navigation instructions \cite{fukushima2022object, wang2023gridmm}. This has led to the incorporation of memory systems into LLMs, such as episodic memory for past experiences and scene graph memory for structured
environmental data. However, because these memory systems often represent memories using natural language, which lacks geometric information, the spatial reasoning capacity of LLMs is compromised.

Alongside these advancements, there is growing interest in using images as a primary source of guidance by integrating foundation models with low-level planners \cite{cai2024bridging, nasiriany2024pivot}. This approach takes advantage of the advanced visual and language understanding capacities of foundation models, offering an effective alternative to traditional map-based methods, which often rely on costly and disruption-prone depth sensing and localization. However, these methods predominantly rely on a first-person perspective without incorporating global memory into the decision-making process. As a result, they often lead to redundant exploration and lower efficiency in complex environments.

In this paper, we propose a novel Vision-Language Model (VLM)-based navigation framework that addresses these challenges by adaptively retrieving task-relevant cues from a global memory module, which are then mapped to the agent’s ego-view visual observations. By integrating global contextual information with local perceptual inputs, our framework enables more informed action decisions, enhancing the agent’s situational awareness and decision-making capabilities. The approach substantially improves the agent’s ability to navigate complex, long-horizon tasks by dynamically aligning global context with egocentric reasoning, providing a more effective and scalable solution for embodied navigation. The experimental results demonstrate that our proposed navigation pipeline outperforms state-of-the-art baselines on both the HSSD and HM3D benchmarks and works effectively on a real robot. Through an ablation study, we verified the essential nature of each component of our method. Using our proposed data collection approach, the supervised fine-tuned Llama3.2-11B \cite{touvron2023llama} model exhibited superior performance compared to both the vanilla Llama3.2-11B model and GPT-4o \cite{hurst2024gpt}.

\section{Related Work} \label{related}
Existing studies that leverage VLMs and LLMs for navigation can be categorized into the following directions.

\subsection{LLM-based Navigation}
These approaches often construct a global memory map based on image observations and use natural language to describe candidate points for navigation, with action decisions driven by large language models (LLMs).

Several methods fall within this category, including \textbf{LFG} \cite{shah2023navigation}, \textbf{VoroNav} \cite{wu2024voronav}, and \textbf{ESC} \cite{zhou2023esc}. While these methods offer the advantage of maintaining a global map and using high-level reasoning, they also face limitations. The language-based reasoning used for decision-making sacrifices high-dimensional semantic information, such as spatial and geometric details, which can constrain performance in complex environments. Furthermore, translating raw ego-view observations into abstract linguistic descriptions may weaken the model’s capacity for precise spatial reasoning.

% \textbf{openFMNav} \cite{kuang2024openfmnav}

% LFG uses frontier-based exploration and large language models to score potential subgoals and guide navigation based on the robot’s observations and exploration progress. VoroNav introduces Reduced Voronoi Graphs (RVGs) to optimize the robot’s exploration by identifying intersections that provide the best observational opportunities, while the LLM predicts the next best waypoint. ESC uses commonsense knowledge and frontier-based exploration to navigate toward objects in the environment, while openFMNav addresses challenges related to human instructions that imply target objects and zero-shot generalization. These methods employ LLMs to dynamically update a semantic map as the robot explores, enhancing memory and reducing redundant exploration.

\subsection{Value Map-based Navigation}
In this class of methods, a global value function is computed based on ego-view observations, and actions are chosen based on the generated value map instead of using VLMs for decision-making.

Notable approaches in this category include 
\textbf{VLFM} \cite{yokoyama2024vlfm} and \textbf{InstructNav} \cite{long2024instructnav}. VLFM uses a pre-trained vision-language model to generate a language-grounded value map, guiding the agent to explore optimal frontiers. InstructNav extends the idea of goal-directed navigation by introducing a Dynamic Chain of Navigation that breaks down tasks into sequences of actions and landmarks. These methods partially address memory forgetting by integrating global value maps, but they still face challenges. The value map is still constructed based on local observations, and decision-making driven by vision-language models (VLMs) often lacks a comprehensive global perspective. As a result, these approaches frequently lead to suboptimal solutions constrained by local decision-making.

\subsection{VLM-based Navigation}
These approaches directly leverage first-person perspective images as the input of vision-language models (VLMs) to generate action decisions. By using the spatial reasoning capabilities of VLMs, these methods enable the model to interpret complex environmental features from the robot’s current viewpoint, facilitating more informed and context-aware navigation decisions.

\textbf{CoNVOI} \cite{sathyamoorthy2024convoi} and \textbf{PIVOT} \cite{nasiriany2024pivot} exemplify approaches that process first-person images with VLMs to facilitate real-time navigation and decision-making. While effective in leveraging immediate visual inputs, these methods lack mechanisms for incorporating historical observations, often resulting in redundant exploration. This limitation poses challenges in long-horizon tasks, where maintaining contextual awareness of past actions is critical for efficient navigation.
\textbf{VLMNav} \cite{goetting2024end} addresses some of these limitations by integrating both RGB-D images and the robot’s pose information to construct a navigability mask that identifies reachable regions. The model incrementally builds a voxel-based map and refines its action proposals by prioritizing unexplored areas.

% \textbf{NoMaD} \cite{sridhar2024nomad} unifies goal-directed navigation and exploration by using the robot's current image and the goal's image as input. The model includes a transformer backbone for processing visual data and a diffusion model for predicting action sequences. A binary mask is applied to the input to focus on either exploration (excluding the goal) or goal-reaching (including the goal). Despite its innovative design, NoMaD remains constrained by the absence of a global memory, relying solely on the most recent three observations. This limitation restricts its capacity for sustained long-term exploration.

% Recent methods have sought to integrate VLMs more effectively for embodied navigation. \textbf{OpenIN} ~\cite{tang2025openin} focuses on navigation tasks where the robot must locate specific objects that have been moved, introducing a Carrier-Relationship Scene Graph (CRSG) to track objects and their locations. The system uses VLMs to process multimodal instructions and commonsense knowledge to guide navigation decisions.

% \textbf{Uni-NaVid} ~\cite{zhang2024uni} takes a significant step toward unifying different navigation tasks in a single model. It processes both video streams and natural language instructions as input, creating a framework that can generalize across a range of navigation tasks. By training on diverse data, including video question answering and captioning tasks, Uni-NaVid improves its performance in real-world scenarios and enables asynchronous execution for efficiency.

These methods move toward integrating both global and local information more effectively, enabling the robot to navigate complex environments with a better understanding of spatial context. However, challenges remain in optimizing the trade-off between VLMs’ generalization capabilities and the need for precise, real-time navigation.

\section{Method} \label{sec:method}

\subsection{Problem Formulation}
In this work, we focus on the object navigation (ObjNav) task, where an agent begins at a random location within an unseen environment and is tasked with finding and navigating to a target object, denoted by $g$. The agent has no access to a pre-built map and must rely entirely on its sensory inputs for navigation. At each time step $t$, the agent captures an egocentric RGB-D image, denoted by $o^t$, from its onboard RGB-D camera. Additionally, the agent has access to its current location and orientation, which are represented by the extrinsic matrix $M_{\text{ext}}$ of the camera. Using these inputs, the agent must compute and execute a low-level action, $a^t$, that efficiently guides it toward the target object.

\begin{figure*}
    \centering
    \includegraphics[width=\linewidth]{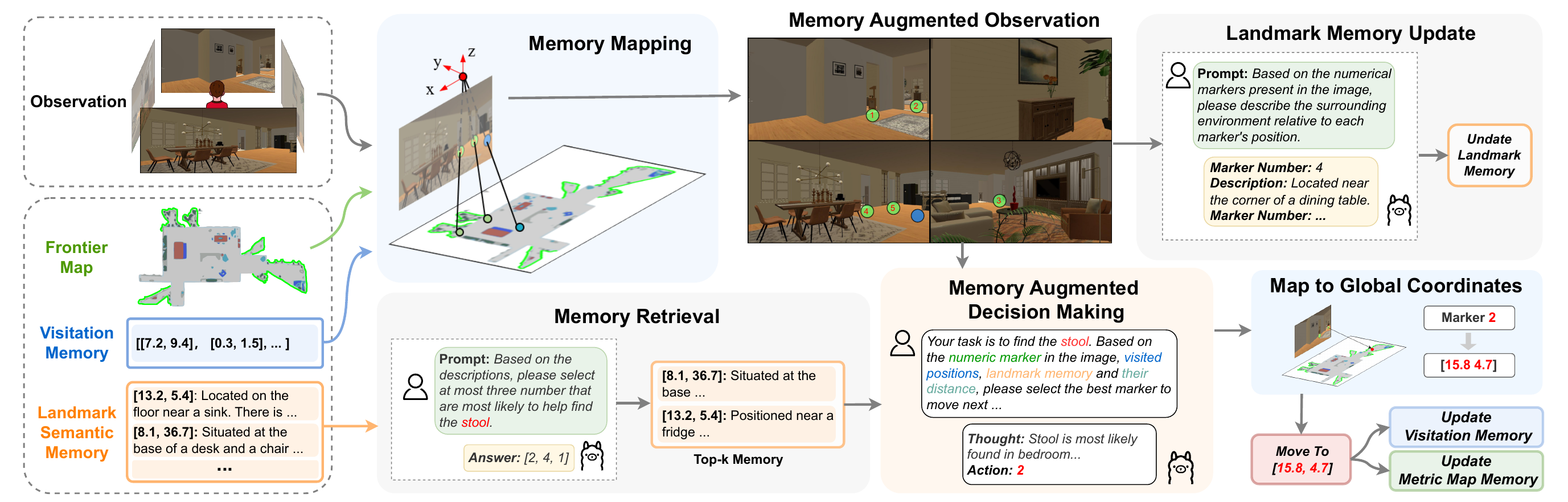}
    \caption{Workflow of our proposed method. Our method maintains three types of memories and project cues from them onto the egocentric images for goal location prediction. Further details are provided in Section~\ref{sec:method}}
    \label{fig:workflow}
\end{figure*}

The workflow of our proposed method is illustrated in Figure~\ref{fig:workflow}. The VLM-based navigation relies on the integration of a memory module that encompasses three distinct types of memories. The construction and maintenance of this memory module, as well as the VLM-based navigation process, will be discussed in detail in the subsequent sections.

\subsection{Memory Construction} \label{sec:method_mem}
The memory module is composed of three distinct types of memories, each serving a different purpose:

\begin{itemize}
    \item \textbf{Frontier Map}: Denoted as $M_f$, frontier map has been proven to be effective for environment exploration in object navigation tasks, as demonstrated in \cite{shah2023navigation, zhou2023esc}. We adopt an approach similar to that used in ESC~\cite{zhou2023esc} to construct the frontier map. Using the agent's position and camera parameters, RGB-D images are transformed into 3D space, where each 2D pixel is mapped to a 3D voxel in the global coordinate system. Voxels located near the floor, with no obstacles along the height dimension, are classified as free space. A frontier in this map is defined as the boundary between free and unexplored areas. This frontier map is maintained throughout the navigation task. 
        
    \item \textbf{Landmark Semantic Memory}: Denoted as $M_l$, this memory stores descriptions of the landmarks that the agent has seen in the past. Each entry includes the global coordinates of the landmark and a description of the nearby semantic information, such as objects or decoration texture. For example: "\textit{[13.2, 5.4]: Located on the floor near a sink. There is a bath tub nearby.}". The description of each landmark is generated by the VLMs, as explained in Section~\ref{sec:method_dec}.

    \item \textbf{Visitation Memory}: Denoted as $M_v$, this memory keeps track of the landmarks that the agent has already visited. 

    By maintaining a record of visited locations, the visitation memory serves as a crucial mechanism to prevent redundant exploration and improve overall exploration efficiency.
\end{itemize}

\subsection{Mem2Ego Navigation} \label{sec:method_dec}
At each time step $t$, given the image-based observation $o^t$ and the three types of memories—$M_f^t$, $M_l^t$, and $M_v^t$—introduced in section~\ref{sec:method_mem}, the proposed memory-to-egocentric (Mem2Ego) navigation process can be formulated as follows:
\begin{equation}
a^t = f_\theta (o^t, M^t_f, M^t_l, M^t_v, g)
\end{equation}

Further details are provided in the following sections. 

\subsubsection{Panoramic Observation Generation}
After the environment is initialized or the agent reaches a new location, the agent captures four egocentric RGB-D images by rotating its viewpoint 90 degrees at each step. These images are then stacked to construct a 360-degree panoramic observation $o_{\text{pano}}^t$ (see Equation~\ref{equation:obs-generation}), offering a comprehensive representation of the surrounding environment. Compared to navigation methods relying on a single egocentric view, this panoramic approach enhances the agent’s spatial awareness and scene understanding. A similar strategy has been employed in \cite{long2024instructnav}.
\begin{equation}
    o_{\text{pano}}^t = \text{Concatenate}([o_{0}^t, o_{\pi /2}^t, o_{\pi}^t, o_{3\pi /2}^t])
    \label{equation:obs-generation}
\end{equation}

\subsubsection{Frontier and Visitation Memory Projection}
Based on the agent's position and the newly captured depth images, the navigation map and corresponding frontiers are updated following the method outlined in Section~\ref{sec:method_mem}. Candidate locations, denoted as $[\mathbf{C}_1,...,\mathbf{C}_N]$ in Equation~\ref{equation:candidates-generation}, are generated by combining frontier clustering and grid-based sampling. The centroid of each frontier segment is computed by clustering all points within the segment. However, using the centroid directly as a candidate may result in unreachable goal positions. To mitigate this, we identify the nearest grid point on the floor area to the centroid, ensuring that the candidate is accessible to the agent. Similarly, visited locations, $[\mathbf{V}_1,...,\mathbf{V}_M]$, are extracted from the visitation memory $M^t_v$, as shown in Equation~\ref{equation:visitation-extraction}. 
\begin{equation} 
    [\mathbf{C}_1,...,\mathbf{C}_N] = \text{CandidatesGeneration}(M_f^t)
    \label{equation:candidates-generation}
\end{equation}
\begin{equation}
    [\mathbf{V}_1,...,\mathbf{V}_M] = \text{VisitationExtraction}(M_v^t)
    \label{equation:visitation-extraction}
\end{equation}
Once determined, the global coordinates of these candidates and visitations are projected onto the egocentric image plane as pixel locations $[\mathbf{c}_1,...,\mathbf{c}_N]$ and $[\mathbf{v}_1,...,\mathbf{v}_M]$, as shown in Equation~\ref{equation:projection}, where $K$ and $M_{\text{ext}}$ represent the camera intrinsics and extrinsics, respectively.
\begin{equation}
\begin{aligned}
    \mathbf{c}_i=\text{Projection}(&\mathbf{C}_i),\quad \mathbf{v}_i=\text{Projection}(\mathbf{V}_i) \\
     \text{where}\hspace{0.1cm} \mathbf{c}_i=(x_i, y_i), \hspace{0.1cm} \mathbf{C}_i=& (X_i, Y_i, Z_i),\hspace{0.1cm}\text{similar for $\mathbf{v}_i$ and $\mathbf{V}_i$} \\
     [x_i',y_i',w_i]^T &= K\cdot M_{\text{ext}}\cdot [X_i,Y_i,Z_i,1]^T \\
     (x_i,y_i)&=(\frac{x_i'}{w_i},\frac{y_i'}{w_i}) 
\end{aligned}
\label{equation:projection}
\end{equation}
An annotation function is then applied to map these locations onto the panoramic observation $o_{\text{pano}}^t$, resulting in an annotated observation $o_{\text{anno}}^t$, as outlined in  Equation~\ref{equation:annotate}. In the annotated image, candidate locations are depicted as green circles, each labeled with a unique identifier corresponding to its position in the image. Similarly, visited locations are marked as blue circles, but only if they are visible within the current view. 
\begin{equation}
    o_{\text{anno}}^t = \text{AnnotateImage}(o_{\text{pano}}^t, [\mathbf{c}_1,...,\mathbf{c}_N], [\mathbf{v}_1,...,\mathbf{v}_M])
\label{equation:annotate}
\end{equation}

\subsubsection{Landmark Memory Retrieval}
The panoramic image, augmented with frontier candidates, highlights potential navigation targets within the agent's immediate field of view. However, it is common that no suitable targets are visible, and more promising options may exist among the landmarks the agent has previously encountered but not yet explored. These previously encountered landmarks are stored in the dynamic landmark semantic memory $M_l^t$. To manage the rapid expansion of this memory during navigation, we utilize large language models (LLMs) to retrieve the top-$k$ landmarks most relevant to the target object. This retrieval process generates an additional observation from memory, $o_{\text{mem}}^t$, which is then incorporated into the decision-making process. % The prompt used for memory retrieval is detailed in Appendix~\ref{sec:infer_prompt}.

\begin{equation}
    o_{\text{mem}}^t = \text{MemoryRetrieval}_{\text{LLMs}}(M_l^t, k)
\end{equation}

\subsubsection{Memory Augmented Decision Making} \label{sec:marker_sel}
At this stage, the panoramic image with annotations, $o_{\text{anno}}^t$, along with the top-$k$ landmarks retrieved from memory, $o_{\text{mem}}^t$, is used to query VLMs to select the next target location to visit (described in Equation~\ref{equation:decision-making}). The VLMs are tasked with identifying the marker on the image most likely to lead to the target object, while avoiding markers that are too close to previously visited locations. To ensure consistency in the output format, the top-$k$ landmarks are numbered, and their descriptions are considered only if no suitable marker is identified directly from the panoramic image. A Chain-of-Thought (CoT) prompting strategy is employed to guide the VLMs in generating a structured reasoning process before producing a single numerical output corresponding to the selected marker. % The full prompt used for decision-making is provided in Appendix~\ref{sec:infer_prompt}.

\begin{equation}
    a^t = f_{\text{VLMs}} \left(\text{prompt}(g), o_{\text{anno}}^t, o_{\text{mem}}^t\right)
\label{equation:decision-making}
\end{equation}

\subsubsection{Action Execution} \label{sec:act_exec}
The marker selected in step~\ref{sec:marker_sel} is transformed to the global coordinate system to determine the global coordinates of the target location. Shortest path follower provided by habitat simulator is then executed to navigate agent to the target location while avoiding obstacles. Object detection is performed each time the agent moves or adjusts its viewing angle. The task is deemed successful if the target object is detected within the agent's field of view and the agent successfully navigates to the target object's viewpoints provided by the Habitat dataset. If the target object is not detected, the process continues until the agent either reaches the designated viewpoints or exceeds the maximum allowed number of exploration steps.

\subsubsection{Memory Update}
While only one landmark from the current view is selected as the next-step navigation target, other landmarks may still be valuable for future exploration. The landmark semantic memory is updated before target position navigation described in Section~\ref{sec:act_exec}. VLMs are prompted to describe the surrounding environment near each marker annotated on the panoramic image. The output from the VLMs includes a list of marker IDs paired with corresponding descriptions. The marker IDs are then converted to global coordinates and, together with their descriptions, saved to the landmark semantic memory for use in future exploration processes. % The prompt used for landmark description is provided in Appendix~\ref{sec:infer_prompt}. 

Meanwhile, the navigation map is updated along the navigation process, using the RGB-D images captured along the way. Additionally, the agent's most recent location is added to the visitation memory to facilitate future exploration.

\subsection{Data Collection and Model Finetuning} \label{sec:method_sft}

\begin{figure}[t]
    \centering
    \includegraphics[width=\linewidth]{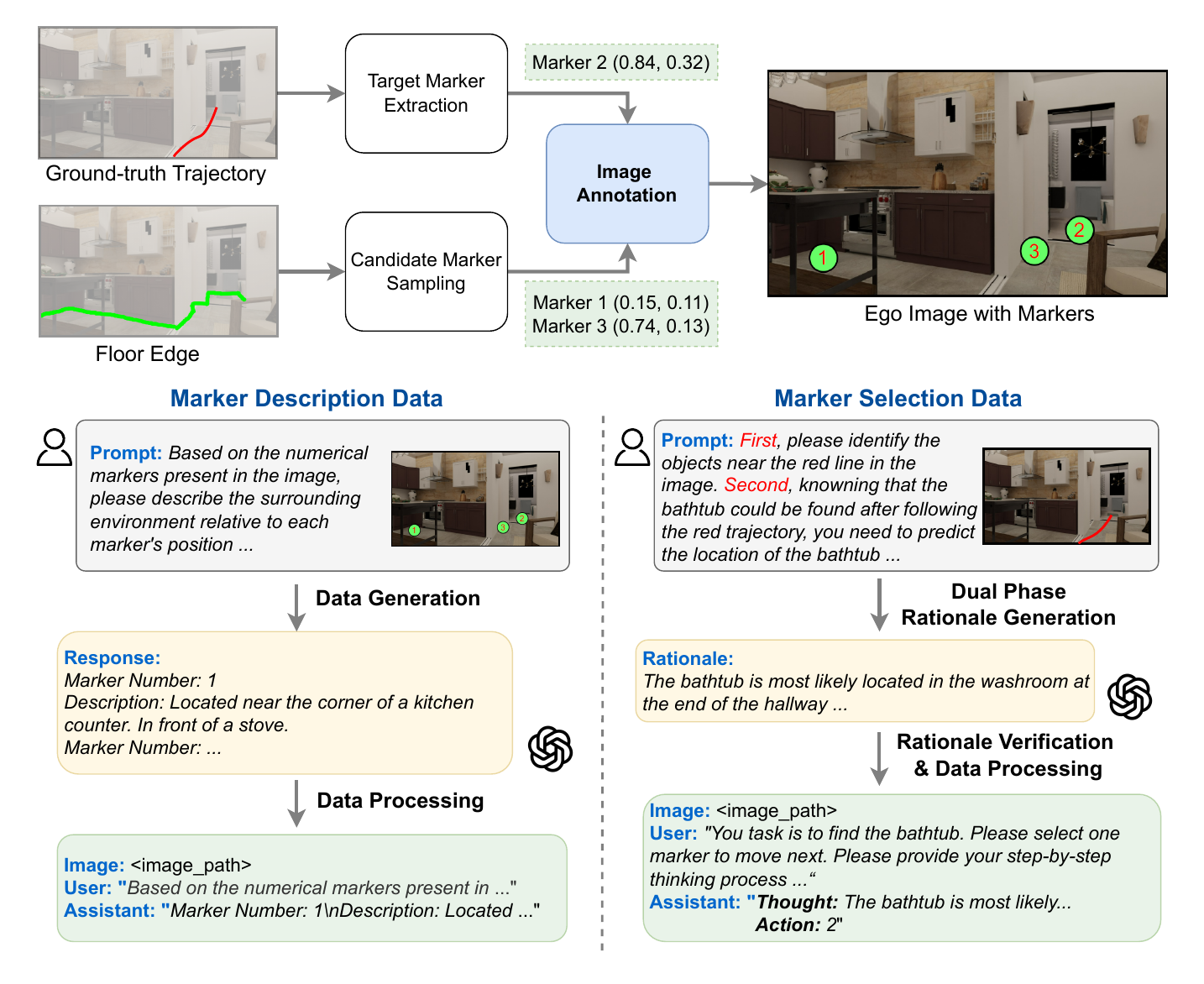}
    \caption{Pipeline of SFT data collection. The ground-truth trajectory and floor edge are used to extract target marker and candidate markers, respectively. Marker description and selection data is generated for model fine-tuning.}
    \label{fig:data_collect}
\end{figure}

To enhance the capabilities of open-sourced VLMs and narrow their performance gap with GPT-4o, we design a pipeline to collect training data for supervised finetuning (SFT). The data collection pipeline is illustrated in Figure~\ref{fig:data_collect}. To improve data diversity and validate the generalization ability of the model, we gather 40 new categories of objects from the HSSD dataset, rather than using the original 6 categories provided. First, new target objects are sampled from the HSSD scenes. For each frame of data, ground-truth trajectories from the current position to these targets are calculated based on the $A^*$ algorithm and subsequently smoothed using Bézier curves. Egocentric images and the corresponding ground-truth target pixel $(x, y)$ (defined as the endpoint of the ground-truth trajectory shown in the image) for each image are saved. To construct the multiple marker annotated image that VLMs encounter in the marker selection task, we generate a few candidate landmarks for each image by sampling from the edge of the floor area. Both ground-truth and sampled candidate landmarks are annotated on the egocentric image in the same way as in the Section~\ref{sec:method}. 

We collect two types of data for VLM fine-tuning: \textbf{marker description} and \textbf{target marker selection with rationale}. To generate marker description data, we use GPT-4o to describe the surrounding environment of each marker on the image. For example, ``\textit{Marker Number: 1 Description: Positioned near a dining chair...; Marker Number: 2 ...}''. Each target marker selection data entry includes both a rationale and the ID of the selected marker. To ensure a robust rationale, we utilizes egocentric images annotated with the ground-truth trajectory and employ a dual-phase prompting strategy: first, GPT-4o is prompted to describe all the objects along the ground-truth trajectory, then to predict the location of the target marker based on its relationship to these objects. Importantly, the rationale generated by GPT-4o must not reference the ground-truth trajectory itself; the trajectory is only used to guide the generation of the rationale. The generated rationale is then automatically validated using GPT-4o, assessing both the accuracy of detected objects and the correctness of the rationale. This dual-phase prompting strategy has proven to be more reliable than a single-phase prompting approach. % The prompts used for rationale generation are provided in Appendix~\ref{sec:sft_data_prompt}.
The validated rationale is then concatenated with the ground-truth marker ID to enforce a CoT-like thinking process. An example of the resulting response is "\textit{Think: The candle is most likely located on the shelf on the right side ... Action: 2}". Note that the resulting marker selection data used for model fine-tuning relies on images annotated with numerical markers, rather than those annotated with the ground-truth trajectory. In total, we generated 30,352 visual question answer (VQA) pairs of data from 104 scenes and 5678 object navigation tasks. This data was used to fine-tune a Llama3.2-11B-Vision model~\cite{grattafiori2024llama3herdmodels} following the configuration recommended by official Llama repository. The model was fine-tuned for 3 epochs with a learning rate of 1e-5 and an effective batch size of 128. 

\section{Experiments}

\subsection{Experimental Setup} \label{sec:exp_setup}

We evaluated our method on the navigation tasks using the Habitat 3.0~\cite{puig2023habitat} simulation platform. We adopt similar setup as the Habitat ObjectNav 2022 challenge~\cite{habitatchallenge2022} for all the experiments. The action space of the agent consists of: \texttt{STOP}, \texttt{MOVE\_FORWARD}, \texttt{TURN\_LEFT}, \texttt{TURN\_RIGHT}, with a forward movement distance of 0.25 meters and a turning angle of 30 degrees per step. For low-level movement control, we utilized Habitat’s built-in shortest-path follower. The maximum number of steps allowed per task is set to 500 by default. Due to limitations in the image quality within the Habitat environment and the suboptimal performance of state-of-the-art perception modules, such as GroundingDINO~\cite{liu2025grounding}, we opted for Habitat’s built-in semantic ground truth with object size conditions as the perception module.  In this context, we can assume that the perception module is sufficiently effective. The LLMs and VLMs used in this study was GPT-4o and Llama3.2-11B.

\subsection{Datasets}

Our method is evaluated on the following three object navigation datasets:

\begin{itemize}
    \item \textbf{Habitat Synthetic Scenes Dataset (HSSD)~\cite{khanna2024habitat}:} We use the HSSD validation dataset to evaluate our method. HSSD consists of 41 scenes and six object goal categories: \texttt{chair}, \texttt{couch}, \texttt{potted plant}, \texttt{bed}, \texttt{toilet}, and \texttt{tv}. To ensure task diversity, we select only one episode per scene-object pair. After filtering out erroneous episodes—such as cases where the agent's initial position was in mid-air—the final number of evaluated episodes is 213.
    
    \item \textbf{HSSD-Hard:} Since some HSSD episodes are relatively easy, with the agent finding the target object in just a few steps, we created a more challenging dataset, HSSD-Hard, by selecting HSSD episodes with longer search distance. We calculated the geodesic distance from the agent's starting point to the target object for each episode and selected the top 50\% of episodes with the longest searching distances to form the HSSD-Hard dataset.  The total number of episodes in HSSD-Hard is 102.

    \item \textbf{HM3D~\cite{ramakrishnan2021habitat}:} We utilized the validation set from the HM3D benchmark, which consists of a total of 1,000 episodes, to evaluate the performance of our method.
    
\end{itemize}

\subsection{Baselines}
We compare our method against the following state-of-the-art (SOTA) baselines that represent different strategies to address the object navigation problem:
\begin{itemize}
    \item \textbf{PIVOT~\cite{nasiriany2024pivot}:} This approach casts the navigation task as an iterative visual question answering problem by annotating the image with numerical markers that represent the navigation sub-goals. The method is adapted for the HSSD object navigation tasks. Without a frontier map, visitation memory, and landmark semantic memory, our proposed navigation pipeline degenerates to PIVOT.
    \item \textbf{LFG~\cite{shah2023navigation}:} This method employs frontier-based exploration and LLMs to score potential sub-goals and guide the navigation.
    \item \textbf{VLFM~\cite{yokoyama2024vlfm}:} The approach utilizes VLMs to generate a language-grounded value map,  from which the location with the highest value is selected as the next sub-goal for navigation.
    \item \textbf{InstructNav~\cite{long2024instructnav}:} InstructNav introduces a Dynamic Chain of Navigation, breaking down navigation tasks into sequences of actions and landmarks. It employs four value maps, each with different semantic representations, to assist in selecting the appropriate landmark. Our implementation of InstructNav differs slightly from that described in the original paper to ensure a fair comparison with our proposed method. The key differences include two main aspects: (1) snapping waypoints to the navigation mesh to ensure they are navigable; if a waypoint cannot be snapped, we sample a random one instead, and (2) using ground-truth segmentation for object detection. All other components remain the same as in the original implementation.
    \item \textbf{VLMNav~\cite{goetting2024end}:} This approach relies on a voxel map built from RGB-D images and the agent's pose to narrow down action space. 
\end{itemize}
To ensure a fair comparison, all experimental setups are conducted under the same conditions described earlier.

\subsection{Metrics}
We employ the following metrics to evaluate the performance of all the methods:
\begin{itemize}
    \item \textbf{Success Rate (SR):} A task was deemed successful when the distance between the agent and any viewpoint of the target object was less than 0.2 meters.
    \item \textbf{Success Weighted by Path Length (SPL)~\cite{anderson2018evaluation}:} This metric evaluates how efficient the agent's path is compared to the optimal path. SPL is calculated as:
    \begin{equation}
        SPL = \frac{1}{N}\sum_{i=1}^N S_i \frac{l_i}{max(p_i, l_i)}
    \end{equation}
    where $l_i$ is the length of the optimal path for episode $i$. $p_i$ is the length of path taken by the agent. $S_i$ is the binary indicator of success in episode $i$.
\end{itemize}

\subsection{Main Results}

% \begin{table}[t]
% \caption{Main results. Our proposed method is compared to the baselines on HSSD and HSSD-Hard datasets. SR: Success Rate. SPL: Success Weighted by Path Length. All experiments are conducted using gpt-4o.}\label{tab:main}
% \centering
% \begin{tabular}{@{}lllll@{}}
% \toprule
%             & \multicolumn{2}{c}{HSSD} & \multicolumn{2}{c}{HSSD-Hard} \\ 
%             \cmidrule(lr){2-3}\cmidrule(lr){4-5}
%             & SR $\uparrow$ & SPL $\uparrow$ & SR $\uparrow$ & SPL $\uparrow$ \\ \midrule
% LFG         & 0.6244       & 0.3371      & 0.6176         & 0.3454         \\
% VLMNav      & 0.6526       & 0.3620      & 0.5294      & 0.1973          \\
% InstructNav-GT & 0.7605       & 0.3722      & 0.6372        &  0.4187       \\
% VLFM        & 0.7652       & 0.5574      & 0.6078         & 0.4270         \\
% PIVOT       & 0.7840       & 0.5658      & 0.6372         & 0.4744         \\
% Ours        & \textbf{0.8685}       & \textbf{0.5788}      & \textbf{0.7647}         & \textbf{0.4790}         \\ \bottomrule
% \end{tabular}
% \end{table}

\begin{table}[t]
\caption{Main results. Our proposed method is compared to the baselines on HSSD, HSSD-Hard and HM3D datasets. SR: Success Rate. SPL: Success Weighted by Path Length. Experiments are conducted using GPT-4o. The HM3D baseline results are sourced from the original papers, with the exception of the PIVOT results, which are from our implementation.}\label{tab:main}
\centering
\begin{tabular}{@{}lllllll@{}}
\toprule
            & \multicolumn{2}{c}{HSSD} & \multicolumn{2}{c}{HSSD-Hard}
            & \multicolumn{2}{c}{HM3D-v2}\\ 
            \cmidrule(lr){2-3}\cmidrule(lr){4-5}\cmidrule(lr){6-7}
            & SR $\uparrow$ & SPL $\uparrow$ & SR $\uparrow$ & SPL $\uparrow$ & SR $\uparrow$ & SPL $\uparrow$ \\ \midrule
LFG         & 0.6244       & 0.3371      & 0.6176         & 0.3454    & 0.689 & 0.36     \\
VLMNav      & 0.6526       & 0.3620      & 0.5294      & 0.1973     & 0.504 & 0.21     \\
InstructNav & 0.7605       & 0.3722      & 0.6372        &  0.4187   & 0.58 & 0.209    \\
VLFM        & 0.7652       & 0.5574      & 0.6078         & 0.4270     & 0.636 & 0.325    \\
PIVOT       & 0.7840       & 0.5658      & 0.6372         & 0.4744     & 0.531 & 0.324   \\
Ours        & \textbf{0.8685}       & \textbf{0.5788}      & \textbf{0.7647}         & \textbf{0.4790}    & \textbf{0.766} & \textbf{0.441}     \\ \bottomrule
\end{tabular}
\end{table}

We evaluate our method and all baselines on both HSSD and HSSD-Hard datasets using the same setup described in Section~\ref{sec:exp_setup}, with results summarized in Table~\ref{tab:main}. Performance is evaluated based on Success Rate (SR) and Success weighted by Path Length (SPL). While SR indicates the overall ability to find the target object, SPL measures the efficiency of the navigation process. Notably, these two metrics are not correlated, as a method can achieve a high SR by sacrificing navigation efficiency.  As shown in Table~\ref{tab:main}, on the HSSD dataset, our proposed method achieves an SR of 0.8685 and an SPL of 0.5788, both of which are higher than all the baseline methods. On the HM3D benchmark, our proposed method also demonstrates superior performance compared to the baselines, achieving an SR of 0.766 and an SPL of 0.441. The results indicate the efficiency of our proposed method in diverse complex environments.

Compared to HSSD, tasks in the HSSD-Hard dataset are more challenging due to the relatively longer search distance, requiring additional steps to locate target objects. As shown in Table~\ref{tab:main}, the performance of all methods decreases on the HSSD-Hard dataset, though the impact varies by model. Notably, our method demonstrate an even greater advantages in these more difficult scenarios, achieving an SR that is 12.75\% higher than the second-best baseline (PIVOT). Additionally, our method outperforms others in SPL as well, further highlighting its efficiency. These results underscore the effectiveness and robustness of our approach in tackling challenging navigation tasks. 
% We also evaluated the performance of our method using the HM3D benchmark, achieving an SR of 0.7654 and an SPL of 0.4407. The results indicate a significant improvement over other baseline methods.

% \begin{figure}[ht]
%     \centering
%     \includegraphics[width=\linewidth]{figures/episode_gain.png}
%     \caption{Reduction of steps per episode on the HSSD-Hard dataset. For each episode, we compute the difference in the number of steps required between our method and the designated baseline. A positive value (blue color) indicates a reduction in steps when using our method, while a negative value (orange color) signifies an increase. Episode IDs are sorted based on these delta values.}
%     \label{fig:episode_gain}
% \end{figure}

% Figure~\ref{fig:episode_gain} illustrates the difference in the number of steps required between our method and the baselines for each episode. For failed episodes, the step count is set to the maximum allowed limit (500 in our case). As shown in the figure, our method reduces or at least matches the number of steps in most episodes, highlighting its superior efficiency in challenging object navigation tasks.

\begin{figure}[t]
    \centering
    \includegraphics[width=\linewidth]{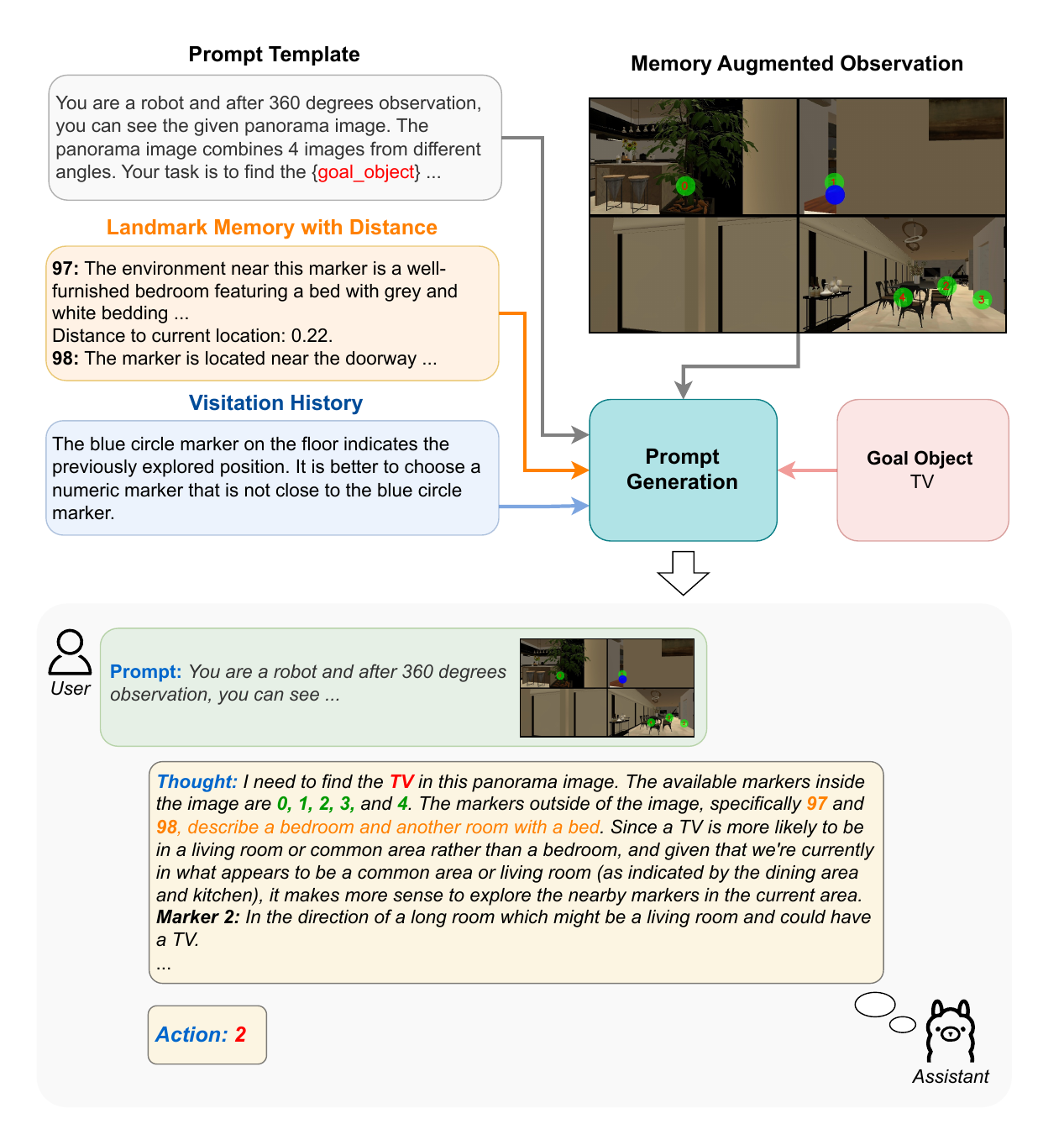}
    \caption{Demonstration of memory-augmented decision making process from a real HSSD episode.}
    \label{fig:good_case_1}
\end{figure}

% The full prompts and GPT-4o response are provided in Appendix~\ref{sec:app_case_study}

Figure~\ref{fig:good_case_1} illustrates the memory-augmented decision-making process from a real HSSD episode. VLMs, such as GPT-4o, analyze all memory cues on the image before reasoning about the most likely location of the target object and selecting the next marker to explore. % The full prompt and responses for this case are provided in Appendix~\ref{sec:app_case_study}. 

Most failed cases with our method result from reaching the maximum allowable number of steps. This can occur due to the VLM selecting a suboptimal position or the task being inherently challenging. Additionally, we occasionally observe that even a strong VLMs like GPT-4o can exhibit visual hallucinations~\cite{liu2024survey}, where they select a marker ID that does not appear in the image or prompt. % We have provided two examples in the Appendix~\ref{sec:visual_hallucinations}.

\begin{table}[h]
\caption{Results with different VLM models. The fine-tuned Llama3.2-11B model (SFT Llama3.2-11B) trained on our collected data outperforms GPT-4o, achieving the best overall performance.} \label{tab:sft_res}
\centering
\begin{tabular}{@{}lllll@{}}
\toprule
            & \multicolumn{2}{c}{HSSD} & \multicolumn{2}{c}{HSSD-Hard} \\ 
            \cmidrule(lr){2-3}\cmidrule(lr){4-5}
            & SR $\uparrow$ & SPL $\uparrow$ & SR $\uparrow$ & SPL $\uparrow$ \\ \midrule
GPT-4o               & 0.8685 & 0.5788 & 0.7647 & 0.4790        \\
Vanilla Llama3.2-11B & 0.7511      & 0.5582     & 0.7352        & 0.4626        \\
SFT Llama3.2-11B      & \textbf{0.8732} & \textbf{0.5995} & \textbf{0.7843} & \textbf{0.5274} \\ \bottomrule
\end{tabular}
\end{table}

\subsection{VLM Model Supervised Fine-tuning (SFT)}
To evaluate the impact of the VLM used, we assess the performance of various VLMs within our proposed method. As shown in Table~\ref{tab:sft_res}, the vanilla Llama3.2 model performs significantly worse than GPT-4o. Given the substantial differences in model size and training data, it is not surprising that smaller open-source VLMs like Llama3.2-11B underperform compared to a strong proprietary models such as GPT-4o. Failure analysis reveals that Llama3.2-11B is more prone to visual hallucinations and struggles with instruction following, particularly in marker selection and description tasks. This may stem from a lack of relevant training data for Llama3.2-11B, limiting its ability to generalize effectively in these scenarios. 

To improve the performance of VLMs, we collected over 30,000 VQA samples by generating data from 40 new object categories within the HSSD dataset. Rationales generated using a dual-phase prompting strategy are used to guide the marker selection process. We then fine-tuned Llama3.2-11B following the approach detailed in Section~\ref{sec:method_sft}. As shown in Table~\ref{tab:sft_res}, the performance of Llama3.2 is improved significantly after supervised fine-tuning (SFT), surpassing GPT-4o in both SR and SPL metrics on the HSSD and HSSD-Hard datasets. These results are particularly promising, considering that Llama3.2-11B is substantially smaller than GPT-4o (11B vs. an estimated 175B) and more cost-effective to train and deploy. This highlights the effectiveness of our data collection strategy and fine-tuning approach. This performance improvement can be attributed to enhanced instruction adherence and a more effective reasoning process grounded in the given environment.

% \begin{figure}[h]
%     \centering
%     \includegraphics[width=\linewidth]{figures/sft_results_v2.png}
%     \caption{Results for different maximum steps. Experiments are conducted on the HSSD dataset. While SFT Llama3.2-11B consistently achieves the best performance, its advantage over the other two models is most pronounced at 300 maximum steps.}
%     \label{fig:diff_max_steps}
% \end{figure}

% Numerous factors influence the performance of the navigation pipeline, particularly the maximum number of allowed steps. Consequently, we assessed the impact of this factor on performance. As shown in Figure~\ref{fig:diff_max_steps}, with the maximum number of steps increasing, both the Success Rate (SR) and Success weighted by Path Length (SPL) improve for each model. When the maximum number of steps is relatively low, such as 200 or fewer, all methods exhibit suboptimal performance, with only minor differences between models. This is likely because the limited step count prevents the full exploitation of each model's capabilities. Conversely, when the maximum number of steps is large, such as 500 or more, all models reach their respective performance limits, making the differences between them less distinct. It is noteworthy that the performance gap between different models becomes more apparent when the maximum number of steps is set to 300 or 400.

\subsection{Ablation Study}

To assess the contribution of each component in our proposed method, we conducted an ablation study by evaluating its performance without landmark semantic memory or visitation memory. As shown in Table~\ref{tab:ablation_study}, removing either of these memory modules leads to a decline in performance. The frontier map could not be excluded from this analysis, as it is essential for marker generation. The absence of visitation memory results in redundant exploration in some cases, thus reducing both the success rate and navigation efficiency. Meanwhile, without landmark semantic memory, the agent is unable to select a navigation goal globally when no suitable marker is present in its current view, which harms the performance as well. These findings highlight the crucial role of both memory modules in fully leveraging the potential of our proposed method.

\begin{table}[h]
\caption{Ablation study. Removing any of the main components of our design leads to degraded performance on HSSD dataset.} \label{tab:ablation_study}
\centering
\begin{tabular}{@{}lllll@{}}
\toprule
& \multicolumn{2}{c}{HSSD} & \multicolumn{2}{c}{HSSD-Hard} \\ 
\cmidrule(lr){2-3}\cmidrule(lr){4-5}
& SR $\uparrow$ & SPL $\uparrow$ & SR $\uparrow$ & SPL $\uparrow$ \\ \midrule
Ours & \textbf{0.8685} & \textbf{0.5788} & \textbf{0.7647} & 0.4790 \\
Ours w/o Visitation Memory & 0.8450 & 0.5761 & 0.7450 & \textbf{0.4961} \\
Ours w/o Landmark Memory & 0.8356 & 0.5669 & 0.7352 & 0.4795 \\ \bottomrule
\end{tabular}
\end{table}

% \section{Experiments on HM3D}

% We also assessed the performance of Mem2Ego using the HM3D benchmark \cite{ramakrishnan2021habitat}. The results indicate that it significantly outperforms other LLM-based baselines.

% \begin{table}[t]
% \caption{Comparison between our method and other baselines on HM3D benchmark. SR: Success Rate. SPL: Success Weighted by Path Length. The baseline results are sourced from the original papers.}\label{tab:hm3d}
% \centering
% \begin{tabular}{@{}lll@{}}
% \toprule
%             & \multicolumn{2}{c}{HM3D} \\ 
%             \cmidrule(lr){2-3}
%             & SR $\uparrow$ & SPL $\uparrow$ \\ \midrule
% VLMNav      & 0.504       & 0.21\\
% VLFM        & 0.525       & 0.304\\
% InstructNav & 0.58       & 0.209\\
% LFG         & 0.689       & 0.36\\
% Ours        & \textbf{0.765}       & \textbf{0.441}\\ \bottomrule
% \end{tabular}
% \end{table}

\section{Real-world Experiments}

\begin{figure}
    \centering
    \includegraphics[width=\linewidth]{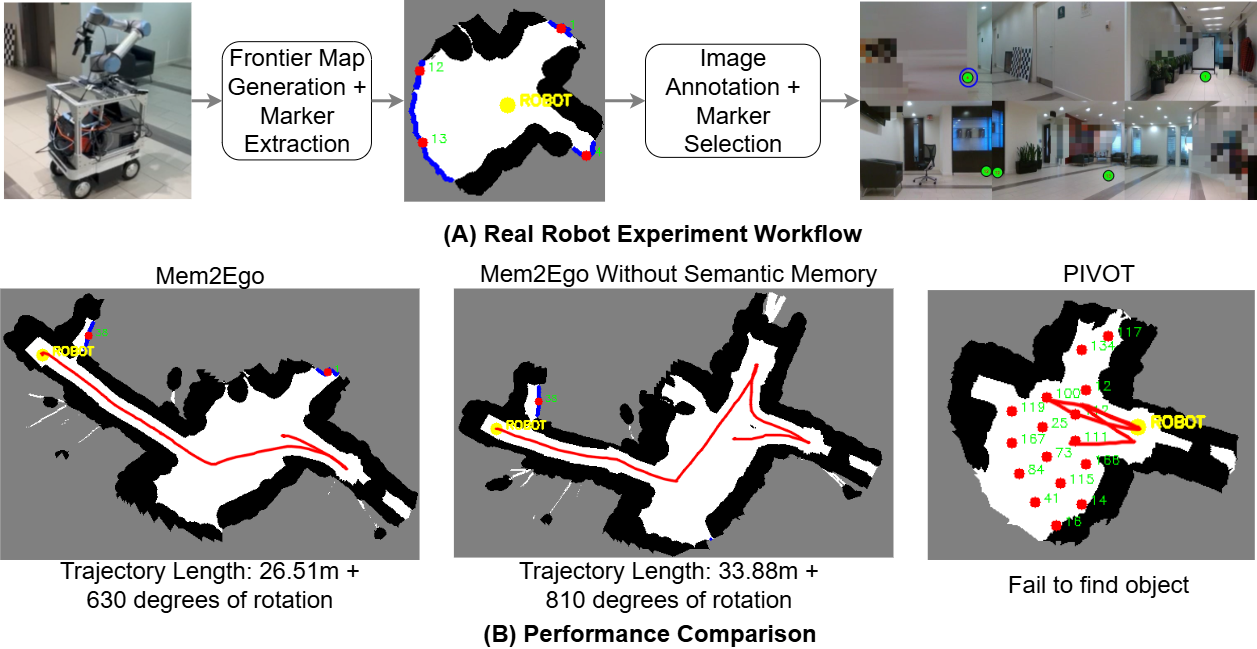}
    \caption{Implementation of Mem2Ego on a real robot. (A) Workflow of the experiment. (B) Performance comparison of Mem2Ego and two baselines.}
    \label{fig:real_robot}
\end{figure}

To evaluate Mem2Ego's performance in a real-world setting, we implemented three methods---Mem2Ego, Mem2Ego without Landmark Semantic Memory, and the baseline method PIVOT \cite{nasiriany2024pivot}---and compared their results. The workflow is illustrated in Figure~\ref{fig:real_robot}(A). The experiments were carried out on a mobile robot platform equipped with three RGB-D cameras. Robot localization relied on cuVSLAM \cite{korovko2025cuvslam} to estimate the base pose. Using the RGB-D data, Nvblox \cite{millane2024nvblox} incrementally generated a 3D reconstruction of the environment, which was further projected into a 2D cost map to support navigation and frontier extraction. The ROS 2 Navigation Stack Nav2 \cite{macenski2020marathon2} used the cost map to plan and execute base motions. The objective of our task is to locate a fire extinguisher in an office environment. The robot initiates the task from a starting position without any prior information about the environment. During exploration, the map and associated memory are incrementally constructed. We employed YOLO-World \cite{cheng2024yolo} as the object detection module, operating at a frequency of 1 Hz.

As shown in Figure~\ref{fig:real_robot}(B), Mem2Ego effectively located the target, with a movement distance of 26.51 meters and a total rotational angle of 630 degrees. In contrast, Mem2Ego without Landmark Semantic Memory eventually located the target after more exploration, resulting in a total movement distance of 33.88 meters and a cumulative rotational angle of 810 degrees. As for PIVOT, it repeatedly moved back and forth within the same area and ultimately failed to locate the target within the 4-minute time limit. These findings are consistent with the simulation.

\section{Limitations}

In our current research, we utilize the VLM to characterize the surrounding environment of each marker and subsequently store the textual response in the landmark semantic memory. This approach heavily relies on the spatial understanding and reasoning capabilities of the VLM, and may result in the potential loss of important semantic information. Additionally, it is worthwhile to explore methods for storing egocentric images in memory and employing the VLM to efficiently process multiple egocentric images.

\section{Conclusion}

This study proposes an efficient fusion strategy that integrates task-relevant global memory information with first-person perspective information, thereby overcoming the suboptimal solution problem associated with existing multimodal navigation frameworks due to local observability. Moreover, this method can simultaneously activate and utilize the complex spatial understanding, reasoning, and commonsense reasoning capabilities of VLMs, thus significantly enhancing the ability and efficiency of navigation decisions in complex spatial scenarios. Theoretically, enhanced spatial cognitive abilities can reduce the required travel distance and number of actions, thereby increasing the task completion success rate and overall efficiency of the navigation scheme.

\bibliographystyle{ieeetran}

\bibliography{root}

\end{document}